\documentclass{article}

\usepackage{arxiv}

\usepackage[utf8]{inputenc} 
\usepackage[T1]{fontenc}    
\usepackage{hyperref}       
\usepackage{url}            
\usepackage{booktabs}       
\usepackage{amsfonts}       
\usepackage{nicefrac}       
\usepackage{microtype}      
\usepackage{lipsum}
\usepackage{graphicx}
\usepackage{bm}
\usepackage{amsmath}
\graphicspath{ {./images/} }
\usepackage{algorithmic}
\usepackage[linesnumbered,lined,boxed,commentsnumbered,ruled,longend]{algorithm2e}
\usepackage{comment}
\usepackage{xcolor}

\newcommand{\argmin}{\mathop{\rm arg~min}\limits}

\title{Federated Rule Ensemble Method in Medical Data}

\author{
    Ke Wan \\
	Department of Medical Data Science\\
	Wakayama Medical University\\
    \texttt{wane19911017@gmail.com} \\
    \And
	Kensuke Tanioka \\
	Department of Biomedical Sciences and Informatics\\
	Doshisha University\\
	\texttt{ktanioka@mail.doshisha.ac.jp} \\
	\And
	Toshio Shimokawa \\
	Department of Biostatistics\\
	Wakayama Medical University\\
	\texttt{toshibow2000@gmail.com} \\
}

\begin{document}
\maketitle

\begin{abstract}
Machine learning has become integral to medical research and is increasingly applied in clinical settings to support diagnosis and decision-making; however, its effectiveness depends on access to large, diverse datasets, which are limited within single institutions. Although integrating data across institutions can address this limitation, privacy regulations and data ownership constraints hinder these efforts. Federated learning enables collaborative model training without sharing raw data; however, most methods rely on complex architectures that lack interpretability, limiting clinical applicability. Therefore, we proposed a federated RuleFit framework to construct a unified and interpretable global model for distributed environments. It integrates three components: preprocessing based on differentially private histograms to estimate shared cutoff values, enabling consistent rule definitions and reducing heterogeneity across clients; local rule generation using gradient boosting decision trees with shared cutoffs; and coefficient estimation via $\ell_1$-regularized optimization using a Federated Dual Averaging algorithm for sparse and consistent variable selection. In simulation studies, the proposed method achieved a performance comparable to that of centralized RuleFit while outperforming existing federated approaches. Real-world analysis demonstrated its ability to provide interpretable insights with competitive predictive accuracy. Therefore, the proposed framework offers a practical and effective solution for interpretable and reliable modeling in federated learning environments.

\end{abstract}

\keywords{Federated Learning \and Interpretable Machine Learning \and Rule Ensemble Model \and Multi-institutional Data Analysis \and Binary outcome}

\section{Introduction}

Over the past decade, machine learning (ML) has emerged as a key tool in medical research, with increasing use in clinical diagnoses and decision-making \cite{Topol2019,Rajkomar2019,Sahni2023,Nguyen2025}. However, its success heavily relies on the availability of large, diverse datasets. In the real world, clinical data are often fragmented across institutions, and datasets at individual sites are limited in size and diversity. This fragmentation can lead to reduced predictive accuracy and poor generalizability, particularly with population heterogeneity and sampling bias \cite{Rieke2020,Sheller2020}. Although integrating data across institutions can alleviate these limitations, the increasingly stringent privacy regulations and data ownership constraints substantially hinder the sharing of sensitive patient-level data \cite{VanPanhuis2014,Tresp2016,Chen2020}.

Federated learning (FL) has emerged as a promising approach to address this challenge, enabling multiple institutions to train models collaboratively without sharing raw data \cite{Mcmahan2017}. Recently, its application in medical research has expanded rapidly, with multiple studies demonstrating its effectiveness in multi-institutional modeling \cite{Brisimi2018,Dayan2021,VAID2021}. However, most existing FL studies have primarily emphasized on predictive performance and relied on complex model architectures, such as deep neural networks \cite{Teo2024,Ma2025}. Consequently, these models often operated as black boxes, offering limited interpretability. This lack of transparency presented a critical limitation in clinical applications, where model interpretability is essential for informed decision-making \cite{Daz-Rodrguez2023,Price2018,Petch2022}.

Interpretable ML has gained increasing attention in recent times \cite{Guidotti2019}. However, a lack of interpretability has been observed in FL settings. Wang \cite{Wang2019} proposed a Shapley value\cite{Shapley1953}-based method to explain feature importance in federated settings while preserving data privacy, and Li et al. \cite{Li2023} discussed the use of post hoc explanation methods, such as SHAP (SHapley Additive exPlanation) and LIME (Local Interpretable Model-agnostic Explanations). However, these approaches provide explanations only after model training and do not fully capture the internal generation of predictions. An alternative is to employ inherently interpretable models. Rule-based methods are particularly preferred in clinical settings because they provide explicit decision rules that align with clinical reasoning \cite{Heerink2023,Xu2023}. Argente et al. \cite{Argente-Garrido2025} proposed a federated decision tree aggregation method that constructs a global interpretable tree through merging decision paths from locally trained trees. However, such models may be prone to instability and often exhibit limited predictive performance compared with more flexible ensemble-based methods. While post-hoc explanation methods permit the use of highly flexible and complex predictive models, inherently interpretable approaches typically rely on simpler model structures, which may limit predictive performance.

RuleFit \cite{Friedman2008} offers a promising solution to this challenge by combining rule-based representations with ensemble learning, achieving strong predictive performance and interpretability. However, its use within FL has not yet been explored. In the conventional RuleFit, the model is constructed in two stages: rule generation and rule ensemble. In the first stage, the rule terms are derived from gradient-boosting decision trees (GBDT; \cite{Friedman2001}). In the second stage, the linear terms are incorporated, and the coefficients of both the rule and linear terms are estimated using LASSO regularization. Extending this framework to FL settings is nontrivial, as it requires performing both stages without sharing raw data across clients. Therefore, the key challenge lies in the implementation of rule generation and rule ensemble in a federated manner while preserving both predictive performance and interpretability. 

To address these challenges, we propose a novel three steps framework for constructing RuleFit models in federated learning settings. First, we introduce a preprocessing step based on differentially private histograms (DP-histograms \cite{Dwork2006}) to estimate global candidate cut-off values for each covariate, thereby reducing the variability of rule definitions and controlling the size of the candidate rule set. Second, we adopt a fully decentralized strategy for rule generation, in which each client independently trains gradient boosting decision tree (GBDT) models using its local data to generate rules. This design avoids communication during tree construction and ensures scalability with respect to the number of clients and boosting iterations, while the shared cut-off candidates help maintain structural consistency across locally generated rules. Third, we formulate the estimation of model coefficients as an $\ell_1$-regularized optimization problem in a federated setting. While naive approaches based on local estimation or simple aggregation may fail to achieve globally consistent variable selection, we adopt a Federated Dual Average (FedDA; \cite{Yuan2021}) algorithm to address this issue. This enables the estimation of a sparse and interpretable model while ensuring that variable selection is performed consistently across institutions. Through these components, the proposed framework provides a practical approach for building interpretable and reliable predictive models in multi-institutional studies without requiring access to individual-level data.

The remainder of this paper is organized as follows. Section 2, describes the RuleFit in federated framework and the model and algorithmic formulation of the proposed method. Section 3 presents simulation studies evaluating its performance relative to existing FL and conventional RuleFit. Section 4 demonstrates its application to real-world clinical data. Finally, Section 5 summarized and discusses implications, limitation of the proposed approach.

\section{Methods}\label{sec2}

\subsection{RuleFit for FL} 

We developed a novel three-step framework for constructing RuleFit models in FL settings. First, we introduced a preprocessing step based on differentially private histograms (DP histograms \cite{Dwork2006}) to estimate global candidate cut-off values for each covariate, reducing the variability of rule definitions and controlling the size of the candidate rule set. Second, we adopted a fully decentralized strategy for rule generation, where each client independently trained GBDT models using local data to generate rules. Finally, we formulated the estimation of model coefficients as an $\ell_1$-regularized optimization problem in a federated setting and adopted a Federated Dual Average (FedDA; \cite{Yuan2021}) algorithm to achieve globally consistent variable selection.

When introducing RuleFit \cite{Friedman2008} to the FL framework, we considered a horizontal FL setting with $M$ clients and binary outcomes. Each client $m = 1,2,\ldots,M$ had a local dataset
\[
D^{(m)} = \left\{(y_i^{(m)}, \bm{x}_i^{(m)})\right\}_{i=1}^{N_m},
\]
where $N_m$ denotes the sample size of the $m$th client, $y_i^{(m)} \in \{0,1\}$ denotes the binary outcome, and $\bm{x}_i^{(m)} = (x_{i1}^{(m)}, x_{i2}^{(m)}, \ldots, x_{ip}^{(m)})^\top \in \mathbb{R}^p$ denotes the $p$-dimensional covariate vector.

We defined the RuleFit model globally as follows:
\begin{align*}
F(\bm{x}) = \beta_0 + \sum_{k=1}^K \beta_k r_k(\bm{x}) + \sum_{j=1}^p \beta_j^* l_j(x_j),
\end{align*}
where $r_k(\bm{x})$ denotes the $k$th rule term; $l_j(x_j)$ denotes the $j$th linear term; $K$ denotes the number of rule terms; and $\beta_0$, $\beta_k$, and $\beta_j^*$ are the intercept, coefficients of the rule terms, and coefficients of the linear terms, respectively. 

Although the model was defined globally, its components were constructed from data distributed across clients without sharing raw data. Details of the rules and linear terms are described below.

\paragraph{Rule terms}
The rule terms $r_k : \mathbb{R}^p \to \{0,1\}$ were defined as conjunctions of the indicator functions
\begin{align*}
r_k(\bm{x}) = \prod_{j=1}^p I(x_j \in S_{jk}),
\end{align*}
where $I(\cdot)$ is the indicator function. $S_j^{(m)} \subseteq \mathbb{R}$ denotes the set of possible values of the $j$th covariate at client $m$, which may differ across clients in an FL framework\cite{Gao2022}. We defined global support as follows:
\[
S_j = \bigcup_{m=1}^M S_j^{(m)}.
\]
Subset $S_{jk} \subseteq S_j$ was defined as follows:
\begin{align*}
S_{jk} = \{x_j < c_{jk}\} \quad \text{or} \quad S_{jk} = \{x_j \ge c_{jk}\},
\qquad (c_{jk} \in \mathcal{C}_j),
\end{align*}
where $c_{jk}$ denotes the cutoff value for the $j$th covariate in the $k$th rule and $\mathcal{C}_j$ denotes the set of candidate cutoff values. Practically, rule candidates are generated locally by each client and aggregated to form a global rule set.

\paragraph{Linear terms}
A model based solely on rule terms may inefficiently approximate linear effects. Therefore, linear terms were incorporated to improve model flexibility. Following Friedman’s approach \cite{Friedman2008}, the $j$th covariate was Winsorized as follows:
\begin{align}
l_j(x_{j}) = \min\left(\delta^+_j, \max\left(\delta^-_j, x_{j}\right)\right),
\label{win_term}
\end{align}
where $\delta^+_j$ and $\delta^-_j$ denote the upper and lower truncation thresholds, respectively. These thresholds are defined as the $(1-q)$- and $q$-quantiles of the aggregated distribution over $S_j$. Accordingly, $q = 0.025$ was set \cite{Friedman2008}.

To ensure that the rule and linear terms had a comparable chance of being selected, the linear terms were normalized as follows:
\begin{align*}
l_j(x_{j}) \leftarrow \frac{0.4 \, l_j(x_{j})}{s^*_j}.
\end{align*}
In centralized settings, $s^*_j$ is computed from all the data. However, in FL, raw data cannot be shared. Therefore, we estimated this using the pooled standard deviation across clients:
\begin{align*}
s^*_j = \sqrt{\frac{\sum_{m=1}^{M}(N_m - 1)\left(s^{(m)}_j\right)^2}{\sum_{m=1}^M (N_m - 1)}},
\end{align*}
where $s^{(m)}_j$ denotes the sample standard deviation of the $j$th covariate for client $m$.

The constant $0.4$ corresponds to the average standard deviation of the rule terms under the assumption that the support of each rule
\begin{align}
\varrho_k = \frac{1}{N} \sum_{i = 1}^N r_k(\bm{x}_i), 
\qquad N = \sum_{m=1}^M N_m,
\label{support}
\end{align}
follows a uniform distribution $U(0,1)$ \cite{Friedman2008}.

\subsection{Proposed Algorithm}

The conventional RuleFit algorithm \cite{Friedman2008} employs a two-step procedure. In the first step, rule terms are generated using GBDT \cite{Friedman2001}. In the second step, linear terms are incorporated, and the coefficients of both the rule and linear terms are estimated using LASSO\cite{Tibshirani1996}. However, this framework assumes that all training data are centrally available, which is not applicable to FL settings.

A straightforward extension is to construct models locally for each client and aggregate them, as proposed by Hauschild et al. \cite{Hauschild2022} for random forests. Although this approach avoids modification of the underlying learning algorithm and reduces communication costs, it results in multiple local models rather than a single global model. Consequently, the predictions rely on several independent models, leading to increased model complexity and reduced interpretability.

To address this limitation, we proposed a federated RuleFit framework that constructs a single global model while preserving data privacy. Rule terms were generated locally by each client using the GBDT; however, owing to differences in observed data across clients, the resulting split points may vary, leading to inconsistent rule definitions. This inconsistency increases the number of rule combinations, resulting in a larger aggregated rule set, and substantially increasing the computational burden in the subsequent rule ensemble step. Therefore, we introduced a preprocessing step that estimates a shared set of candidate cutoff values in a privacy-preserving manner, enabling aligned rule definitions across clients. The resulting rules were aggregated to form a global rule set. Linear terms were then incorporated and the coefficients of both the rule and linear terms were estimated through solving an $\ell_1$-regularized optimization problem in a federated manner using the FedDA \cite{Yuan2021} algorithm.

The proposed algorithm consists of the following three steps:
\begin{enumerate}
    \item \textbf{Pre-processing:} Estimates global candidate cut-off values for each covariate in a privacy-preserving manner.
    \item \textbf{Rule generation:} Each client independently generates rule terms using GBDT based on the predefined cut-off candidates, and the rules are aggregated across clients.
    \item \textbf{Rule ensemble:} A global model is constructed through estimating the coefficients of the rule and linear terms via $\ell_1$-regularized optimization using the FedDA algorithm.
\end{enumerate}

\begin{figure}[!t]
  \centering
  \includegraphics[width=\linewidth]{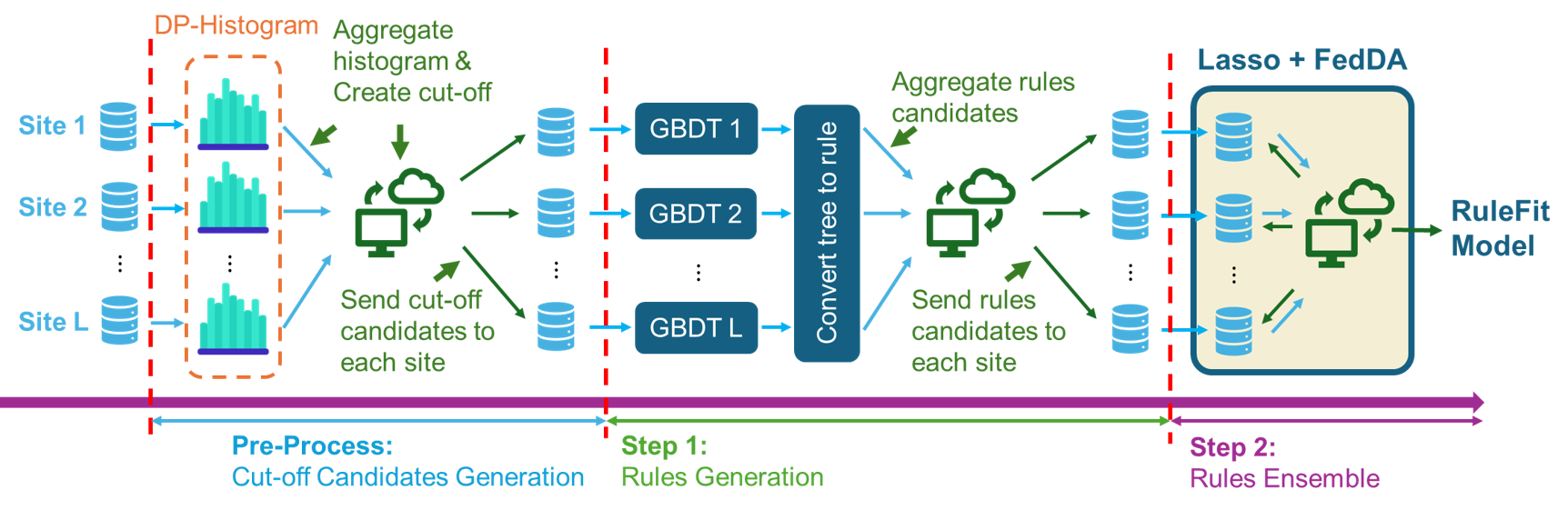}
  \caption{Overview of the proposed federated RuleFit framework}
  \label{Fig1}
\end{figure}

An overview of the proposed algorithm is illustrated in Figure~\ref{Fig1}.

\subsubsection{Pre-processing}

In this step, we constructed a shared set of candidate cut-off values for each covariate to restrict tree splits in the subsequent rule generation step, ensuring consistency across clients. To construct such global cutoff candidates directly, sharing raw data across clients is necessary, which raises privacy concerns. Alternatively, the marginal distribution of each covariate can be approximated through constructing histograms locally for each client, aggregating them at the server, and determining the cut-off candidates based on the resulting quantiles. However, the direct sharing of histogram counts can lead to information leakage \cite{Xu2013}. 

To address this issue, we employed DP histograms \cite{Dwork2006}. In particular, each client perturbs its local histogram count using the Laplace mechanism, ensuring $\epsilon$-differential privacy (Appendix 1). The noisy histograms were aggregated to approximate the global distribution, from which quantiles were derived to define the shared candidate cutoff values. This approach enables privacy-preserving approximation of the global distribution, while retaining sufficient information for quantile-based cut-off selection.

\noindent\textbf{Step 1: Construct DP histograms}

For the $j$th covariate at client $m$, we considered a histogram with $B$ bins defined by the boundaries 
$\tilde{h}_1 < \cdots < \tilde{h}_{B+1}$. 
The bin count was defined as follows:
\begin{align*}
H_b^{(m)} 
=
\sum_{i=1}^{N_m}
I\!\left(x_{ij}^{(m)} \in [\tilde{h}_b, \tilde{h}_{b+1})\right),
\quad b=1,\ldots,B,
\end{align*}
where $\tilde{h}_b$ and $\tilde{h}_{b+1}$ denote the boundaries of the $b$th bin, and $b$ indicates the histogram bins.

To ensure $\epsilon$-differential privacy, Laplace noise with scale $1/\epsilon$ was added independently to each bin count as follows:
\begin{align*}
\tilde{H}_b^{(m)} 
=
H_b^{(m)} 
+
\mathrm{Lap}\!\left(0, \frac{1}{\epsilon}\right),
\end{align*}
where $\mathrm{Lap}(0, 1/\epsilon)$ denotes a Laplace random variable with a mean $0$ and scale parameter $1/\epsilon$.

\noindent\textbf{Step 2: Aggregate histograms across clients}

The server aggregated noisy histograms through summing them across the clients as follows:
\begin{align*}
\tilde{H}_b^{\text{global}}
=
\sum_{m=1}^{M} \tilde{H}_b^{(m)}.
\end{align*}

The cumulative distribution function (CDF) was approximated as follows:
\begin{align*}
\hat{F}(\tilde{h}_b) 
=
\frac{\sum_{l=1}^{b} \tilde{H}_l^{\text{global}}}
{\sum_{l=1}^{B} \tilde{H}_l^{\text{global}}},
\quad b=1,\ldots,B.
\end{align*}

\noindent\textbf{Step 3: Generate candidate cut-off values}

$Q$ denotes the number of quantile levels and defines
\begin{align*}
\tau_q = \frac{q}{Q+1}, 
\quad q=1,\ldots,Q.
\end{align*}

The corresponding quantile estimates were obtained as follows:
\begin{align*}
\hat{q}_{\tau_q}
=
\min \left\{
\tilde{h}_b :
\hat{F}(\tilde{h}_b) \ge \tau_q
\right\}.
\end{align*}

The candidate cutoff for the $j$th covariate was defined as follows:
\begin{align*}
\mathcal{C}_j 
=
\{
\hat{q}_{\tau_1}, \ldots, \hat{q}_{\tau_Q}
\}.
\end{align*}

The collection of candidate sets for all covariates was denoted by
\begin{align*}
\bm{\mathcal{C}} = \{\mathcal{C}_1, \ldots, \mathcal{C}_p\}.
\end{align*}

\subsubsection{Rule Generation}

In this step, we generated rule candidates for the proposed model. Rules were constructed for each client by fitting GBDT models to local data, where split points were restricted to a predefined set of candidate cutoff values obtained during the pre-processing step. The rule sets obtained from all the clients were aggregated to form the final global rule candidate set. The detailed procedure is presented in Algorithm~\ref{alg1}.

\noindent\textbf{Step 1: Fit GBDT models for each client}

For each client, $m = 1,2,\ldots,M$, a GBDT model with $C$ boosting iterations was constructed using local data. At each split, the candidate split points were restricted to the predefined set $\bm{\mathcal{C}}$ obtained in the pre-processing step, ensuring that all tree partitions were defined using the shared cut-off values.

The GBDT model constructed for client $m$ was expressed as follows:  
\begin{align*}
F^{(m)}(\bm{x}) = \sum_{c = 1}^C f^{(m)}_c(\bm{x}),
\end{align*}
where $f_c^{(m)} : \mathbb{R}^p \rightarrow \mathbb{R}$ denotes the base learner at the boosting step $c$, which is constructed using the split points restricted to the candidate set $\bm{\mathcal{C}}$.

Each base learner was written as follows:
\begin{align*}
f^{(m)}_c(\bm{x}) = \sum_{t = 1}^{T^{(m)}_c} \gamma^{(m)}_{ct} I(\bm{x} \in R^{(m)}_{ct}(\bm{\mathcal{C}})),
\end{align*}
where $R_{ct}^{(m)}(\bm{\mathcal{C}})$ denotes the $t$th disjoint partition region induced by splits restricted to $\bm{\mathcal{C}}$, $\gamma_{ct}^{(m)}$ is the corresponding leaf weight, and $T_c^{(m)}$ is the number of terminal nodes in tree $c$.

For the RuleFit \cite{Friedman2008}, the number of terminal nodes was randomly determined as follows:
\begin{align*}
T_c^{(m)} 
=
2 + \lfloor \omega \rfloor,
\quad 
\omega \sim \mathrm{Exponential}\!\left(\frac{1}{\bar{L}-2}\right),
\end{align*}
where $\bar{L}$ denotes the average tree depth, $\lfloor \cdot \rfloor$ denotes the floor function, and $\mathrm{Exponential}(\lambda)$ denotes the exponential distribution with the rate parameter $\lambda = 1/(\bar{L}-2)$. This randomization encouraged shallow trees, while allowing occasional deeper trees, enabling the capture of higher-order interaction effects.

\paragraph{Step 2: Extract rule terms}

For client $m$, the set of base learners 
$\{f_c^{(m)}(\bm{x})\}_{c=1}^{C}$ 
was decomposed into rule terms derived from the internal and terminal nodes of the trees corresponding to the conjunctions of the split conditions along each decision path.

For each tree $c$, we obtained the following set of rule terms:
\begin{align*}
\{ r_{k}^{(m)}(\bm{x}) \}_{k=1}^{K_c^{(m)}},
\end{align*}
where $K_c^{(m)} = 2(T_c^{(m)} - 1)$ denotes the number of rule terms extracted from the tree $f_c^{(m)}(\bm{x})$.

Aggregating across all trees and clients yielded the following global rule candidate set:
\begin{align*}
\mathcal{R}
=
\bigcup_{m=1}^{M}
\bigcup_{c=1}^{C}
\{ r_{k}^{(m)}(\bm{x}) \}_{k=1}^{K_c^{(m)}}.
\end{align*}

After removing duplicate rules—defined as rules that were identical or logically equivalent (e.g., $x_1 > 1 \ \& \ x_2 < 1$, equivalent to $x_2 < 1 \ \& \ x_1 > 1$)—the final rule candidate set was defined as follows:
\begin{align*}
\mathcal{R}^{*} = \{ r_k(\bm{x}) \}_{k=1}^{K},
\end{align*}
where $K = |\mathcal{R}^{*}|$ denotes the total number of candidate rules.

\begin{algorithm}[H]
\small
\caption{Rule Generation}\label{alg1}
\DontPrintSemicolon
\SetAlgoLined
\SetKwBlock{Server}{Server side}{}
\BlankLine

\textbf{Input}\\
Number of trees $C$, average depth $\bar{L}$, learning rate $v$, and cutoff candidates $\bm{\mathcal{C}}$. \\
\textbf{Output}\\
Final global rule candidate set $\mathcal{R}^*$. \;

\BlankLine
\Server{
Broadcast the shared cutoff candidates $\bm{\mathcal{C}}$ to all clients. \;
}

\BlankLine
\ForEach{$m=1,2,\ldots,M$ \tcp*{in parallel}}{
Initialize
\[
F_0^{(m)} \leftarrow \log\left(\frac{\sum_{i=1}^{N_m} y_i^{(m)}}{\sum_{i=1}^{N_m} (1-y_i^{(m)})}\right)
\]

\For{$c=1,2,\ldots,C$}{
Compute pseudo-residuals
\[
r_i^{(m,c)} \leftarrow y_i^{(m)} - \frac{1}{1+\exp\left(-F_{c-1}^{(m)}(\bm{x}_i^{(m)})\right)},
\quad i=1,\ldots,N_m
\]

Sample the number of terminal nodes
$T_c^{(m)} = 2 + \lfloor \omega \rfloor$,
where $\omega \sim \mathrm{Exponential}\!\left(\frac{1}{\bar{L}-2}\right)$.\;

Fit a regression tree $f_c^{(m)}(\bm{x})$ to $\{r_i^{(m,c)}\}_{i=1}^{N_m}$ using split points restricted to $\bm{\mathcal{C}}$ to yield regions $\{R_{ct}^{(m)}\}_{t=1}^{T_c^{(m)}}$. \;

For each region $t=1,\ldots,T_c^{(m)}$, estimate
\[
\hat{\gamma}_{ct}^{(m)}
=
\argmin_{\gamma}
\sum_{\bm{x}_i^{(m)} \in R_{ct}^{(m)}} (r_i^{(m,c)}-\gamma)^2
\]

Update
\[
F_c^{(m)}(\bm{x})
=
F_{c-1}^{(m)}(\bm{x})
+
v\sum_{t=1}^{T_c^{(m)}} \hat{\gamma}_{ct}^{(m)} I(\bm{x}\in R_{ct}^{(m)})
\]
}

Extract the local rule candidate set
\[
\mathcal{R}^{(m)}=\{r_k^{(m)}(\bm{x})\}_{k=1}^{K^{(m)}},
\qquad
K^{(m)}=\sum_{c=1}^{C} 2(T_c^{(m)}-1)
\]

Send $\mathcal{R}^{(m)}$ to the server.\;
}

\BlankLine
\Server{
Aggregate local rule sets:
\[
\mathcal{R}=\bigcup_{m=1}^{M}\mathcal{R}^{(m)}
\]

Remove duplicated rules, including logically equivalent rules, to obtain $\mathcal{R}^*$.\;
}
\end{algorithm}

\subsubsection{Rule ensemble}
At this stage, we aimed to estimate the coefficients of the rules and linear terms in the proposed model. In the conventional RuleFit algorithm, these coefficients are obtained by solving an $\ell_1$-regularized optimization problem (i.e., LASSO) under a centralized setting. However, in an FL framework, client-level data cannot be shared across institutions because of privacy constraints. Therefore, the centralized LASSO estimation procedure is not applicable, and a federated optimization strategy is required to estimate the model parameters without exchanging raw data.

In the proposed model, the coefficients of the rule and linear terms were estimated through minimizing the following objective function:
\begin{align}
\label{obj}
&\min_{\{\beta_k\}_{k=0}^K,\{\beta_j^*\}_{j=1}^p}
f(\{\beta_k\}_{k=0}^K, \{\beta_j^*\}_{j=1}^p)\notag \\
&=
\sum_{m=1}^{M}
\Bigg[
\sum_{i=1}^{N_m}
\left\{
\log\left(1+\exp(\eta_i^{(m)})\right)
-
y_i^{(m)} \eta_i^{(m)}
\right\}
\Bigg]
\notag \\
&\quad
+ \lambda\left(
\sum_{k=1}^K \|\beta_k\|_1
+
\sum_{j=1}^p \|_1\beta_j^*\|_1
\right),
\end{align}
where the linear predictor is defined as follows:
\begin{align*}
\eta_i^{(m)}
=
\beta_0
+
\sum_{k=1}^K \beta_k r_k(\bm{x}_i^{(m)})
+
\sum_{j=1}^p \beta_j^* l_j(x_{ij}^{(m)}),
\end{align*}
where $\lambda > 0$ is the regularization parameter that controls the sparsity, and $\|\cdot\|_1$ denotes the $\ell_1$ norm. This predictor corresponds to an $\ell_1$-regularized logistic regression model that combines rules and linear terms.

To optimize this non-smooth convex objective function in an FL setting, we adopted the FedDA algorithm proposed by Yuan et al.~\cite{Yuan2021}.It is well-suited for composite optimization problems involving non-smooth regularizers- such as the $\ell_1$ penalty. FedDA considers a composite optimization problem as follows:
\begin{align*}
\min_{\bm{w}\in\mathbb{R}^p}\Phi(\bm{w}) = f(\bm{w}) + \psi(\bm{w}) = \sum_{m=1}^M f_m(\bm{w}) + \psi(\bm{w}),
\end{align*}
where $p$ denotes the dimensions of the parameter vector, $M$ denotes the number of clients, $f_m(\cdot)$ denotes the loss function at client $m$, and $\psi(\cdot)$ denotes a possible non-smooth regularizer.

To cast \eqref{obj} into this composite form, we defined client-specific loss as 
\begin{align*}
&f_m(\{\beta_k\}_{k=0}^K, \{\beta_j^*\}_{j=1}^p)\\
&=
\sum_{i=1}^{N_m}
\left\{
\log\left(1+\exp(\eta_i^{(m)})\right)
-
y_i^{(m)}\eta_i^{(m)}
\right\}
\end{align*}
and the global regularizer as an $\ell_1$ penalty,
\begin{align*}
\psi(\{\beta_k\}_{k=0}^K, \{\beta_j^*\}_{j=1}^p)
=
\lambda\left(
\sum_{k=1}^K \|\beta_k\|_1
+
\sum_{j=1}^p \|\beta_j^*\|_1
\right).
\end{align*}

Using these definitions, the objective in \eqref{obj} was expressed as a special case of the FedDA formulation, which enables efficient distributed optimization while preserving sparsity and avoiding direct data sharing. Similar to other FL algorithms, FedDA comprises two components: (1) local updates performed at each client and (2) global aggregation performed at the central server. A detailed implementation of FedDA for the proposed model is presented in Algorithms ~\ref{alg_client} (client update) and ~\ref{alg_server} (server aggregation).

\noindent\textbf{Part 1: Server Update}: Algorithm~\ref{alg_server} describes the server-side update procedure of the Federated Dual Averaging (FedDA) algorithm.

At initialization, the global primal parameter vector was defined as follows:
\begin{align*}
\bm{w}_0 \leftarrow \bm{0},
\end{align*}
where
\[
\bm{w}_0 = (\beta_0^{(0)}, \beta_1^{(0)}, \ldots, \beta_K^{(0)}, \beta_1^{*(0)}, \ldots, \beta_p^{*(0)})^\top
\]
denotes the initial coefficient vector of the proposed model.

The corresponding dual variable was initialized as follows:
\begin{align*}
\bm{z}_0 \leftarrow \nabla h(\bm{w}_0),
\end{align*}
where $h(\cdot)$ denotes a strongly convex distance-generating function. In the Euclidean setting
\begin{align*}
h(\bm{w}) = \frac{1}{2}\|\bm{w}\|_2^2,
\end{align*}
, dual mapping was simplified to $\bm{z}_0 = \bm{w}_0$. The working dual variable wass set at $\bm{z}_r \leftarrow \bm{z}_0$.

The server then performed $R$ rounds of communication. In each $r$ round, the dual variable $\bm{z}_r$ was broadcast to all clients. Each client performed $G$ local dual-averaging updates using local data and returned the dual variable pair $(\bm{z}_{r,0}^m, \bm{z}_{r,G}^m)$.

After receiving updates from all the clients, the server computed the average dual increment as follows:
\begin{align*}
\bm{\Delta}_r = \frac{1}{M}\sum_{m=1}^{M}\left(\bm{z}^m_{r,G} - \bm{z}^m_{r,0}\right).
\end{align*}

Each client updated its dual variables as follows:
\begin{align*}
\bm{z}^m_{r,g+1} \leftarrow \bm{z}^m_{r,g} - \eta_c\, \bm{g}^m_{r,g},
\end{align*}
The difference $\bm{z}_{r,G}^m - \bm{z}_{r,0}^m$ corresponds to the accumulated negative gradients over $G$ local iterations. Therefore, $\bm{\Delta}_r$ represents the average accumulated gradient information in the dual space.

The global dual variable was updated as follows:
\begin{align*}
\bm{z}_{r+1} \leftarrow \bm{z}_r - \eta_s \bm{\Delta}_r,
\end{align*}
where $\eta_s$ denotes server step size.

The cumulative regularization coefficient was updated as follows:
\begin{align*}
\tilde{\eta}_{r+1} \leftarrow \eta_s\,\eta_c\,(r+1)G.
\end{align*}

In this formulation, only dual variables were exchanged and updated during the communication rounds. As client-side updates depend solely on dual variables, recovering the primal variable in each round was unnecessary. Instead, the primal variable was obtained only once after completing all communication rounds.

Specifically, after $R$ rounds, the final parameter vector was recovered through proximal mapping.
\begin{align*}
\bm{w}_{R} \leftarrow \argmin_{\bm{w}}
\left\{
\langle -\bm{z}_{R}, \bm{w}\rangle
+ \tilde{\eta}_{R}\lambda \|\bm{w}\|_1
+ h(\bm{w})
\right\}.
\end{align*}

Finally, the coefficients of the proposed model $(\{\beta_{k}\}_{k=0}^K,\{\beta^*_{j}\}_{j=1}^p)^\top$ were obtained from $\bm{w}_R$.

\begin{algorithm}[h]
\caption{FedDA: ServerUpdate}\label{alg_server}
\DontPrintSemicolon
\KwIn{$R$ (number of communication rounds), 
$M$ (number of clients),
$\eta_s$ (server step size), 
$\eta_c$ (client step size), 
$G$ (number of local iterations)}
\KwOut{$(\{\beta_{k}\}_{k=0}^K,\{\beta^*_{j}\}_{j=1}^p)$ (estimated model coefficients)}

$\bm{w}_0 \leftarrow \bm{0}$\;
$\bm{z}_0 \leftarrow \nabla h(\bm{w}_0)$\;
$\bm{z}_r \leftarrow \bm{z}_0$\;

\For{$r=0$ \KwTo $R-1$}{
    Broadcast $\bm{z}_r$ for all clients\;
    \For{$m=1$ \KwTo $M$}{
        Run Algorithm~\ref{alg_client} on client $m$\;
        Receive $(\bm{z}^m_{r,0}, \bm{z}^m_{r,G})$\;
    }
    $\bm{\Delta}_r \leftarrow \frac{1}{M}\sum_{m=1}^{M}(\bm{z}^m_{r,G} - \bm{z}^m_{r,0})$\;
    $\bm{z}_{r+1} \leftarrow \bm{z}_r - \eta_s\,\bm{\Delta}_r$\;
    $\tilde{\eta}_{r+1} \leftarrow \eta_s\,\eta_c\,(r+1)G$\;
}

$\bm{w}_{R} \leftarrow \argmin_{\bm{w}}\{\langle -\bm{z}_{R}, \bm{w}\rangle + \tilde{\eta}_{R}\lambda \|\bm{w}\|_1 + h(\bm{w})\}$\;

$(\{\beta_{k}\}_{k=0}^K,\{\beta^*_{j}\}_{j=1}^p)^\top \leftarrow \bm{w}_{R}$\;
\Return $(\{\beta_{k}\}_{k=0}^K,\{\beta^*_{j}\}_{j=1}^p)^\top$\;
\end{algorithm}

\noindent\textbf{Part2: Client Update}: Algorithm~\ref{alg_client} describes the local update procedure executed on client $m$ within the Federated Dual Averaging (FedDA) framework.

At the beginning of round $r$, the local dual variable was initialized as follows:
\begin{align*}
\bm{z}_{r,0}^m \leftarrow \bm{z}_r,
\end{align*}
where $\bm{z}_r$ is the global dual variable obtained in the previous communication round.

Then, the client performed $G$ local inner iterations. At each $g$ step, the cumulative step-size parameter was defined as follows:
\begin{align*}
\tilde{\eta}_{r,g} \leftarrow \eta_s\,\eta_c\,rG + \eta_c\,(g+1),
\end{align*}
This reflected the accumulated scaling of the regularization term across both the previous communication rounds and current local iterations, which is essential in the dual-averaging framework, where updates depend on the aggregation of past gradients rather than on only the current one \cite{Nesterov2009,Xiao2010}.

At each local iteration, the primal variable was recovered from the dual variable through proximal mapping.
\begin{align*}
\bm{w}^m_{r,g} \leftarrow \argmin_{\bm{w}\in\mathbb{R}^p}
\left\{
\langle -\bm{z}^m_{r,g}, \bm{w}\rangle 
+ \tilde{\eta}_{r,g}\lambda \|\bm{w}\|_1 
+ h(\bm{w})
\right\}.
\end{align*}

The gradient of the local loss function was then computed as follows:
\begin{align*}
\bm{g}^m_{r,g} \leftarrow \nabla f_m\!\left(\bm{w}^m_{r,g}; D^{(m)}\right),
\end{align*}
and the dual variable is updated as:
\begin{align*}
\bm{z}^m_{r,g+1} \leftarrow \bm{z}^m_{r,g} - \eta_c\, \bm{g}^m_{r,g}.
\end{align*}

After completing $G$ local iterations, the client returned the pair $(\bm{z}_{r,0}^m, \bm{z}_{r,G}^m)$ to the server.

\begin{algorithm}[h]
\caption{FedDA: Client Update (client $m$)}\label{alg_client}
\DontPrintSemicolon
\KwIn{$\bm{z}_r$ (global dual variable), $D^{(m)}$ (local dataset), 
$\eta_c$ (client step size), 
$\eta_s$ (server step size), 
$G$ (number of local iterations)}
\KwOut{$\bm{z}^m_{r,0}$, $\bm{z}^m_{r,G}$ (initial and updated local dual variables)}

$\bm{z}^m_{r,0} \leftarrow \bm{z}_r$\;

\For{$g=0$ \KwTo $G-1$}{
    $\tilde{\eta}_{r,g} \leftarrow \eta_s\,\eta_c\,rG + \eta_c\,(g+1)$\;
    $\bm{w}^m_{r,g} \leftarrow \argmin_{\bm{w}}\{\langle -\bm{z}^m_{r,g}, \bm{w}\rangle + \tilde{\eta}_{r,g}\lambda \|\bm{w}\|_1 + h(\bm{w})\}$\;
    $\bm{g}^m_{r,g} \leftarrow \nabla f_m(\bm{w}^m_{r,g}; D^{(m)})$\;
    $\bm{z}^m_{r,g+1} \leftarrow \bm{z}^m_{r,g} - \eta_c\, \bm{g}^m_{r,g}$\;
}
\Return $\bm{z}^m_{r,0}, \bm{z}^m_{r,G}$\;
\end{algorithm}

\subsection{Interpretation tools}

The proposed approach created an additive model using rule and linear terms as base functions to facilitate the interpretation of the relationship between the covariates and outcomes. While the constructed models included certain rules, in actual applications, we focused on the most important ones. The original RuleFit provides interpretation tools, including base function and rule importance, to evaluate the importance of the base function and variables. In this study, we extended them to FL and provided a detailed description as follows:

\noindent{\textbf{Base function importance}} The base function importance includes the importance of the rule and linear terms. A high or low base-function importance value indicates that the corresponding base function contributed to the outcome. The importance of the rule and linear terms was as follows: 
\begin{align*}
\displaystyle
I_k(\bm{x}) &= \left|\hat{\beta}_k\right|\cdot std_{pool}(r_k(\bm{x})) \\
&= |\hat{\beta}_k|\cdot \sqrt{\frac{\sum_{m=1}^{M}(N_m-1)std(r_k(\bm{x}^{(m)}))}{\sum_{m=1}^M(N_m-1)}} \quad \mathrm{and} \\
I_j(x_{j}) &= \left|\hat{\beta}^{*}_j\right|\cdot std_{pool}\left(l_j(x_{j})\right)\\
& = |\hat{\beta}^*_k|\cdot \sqrt{\frac{\sum_{m=1}^{M}(N_m-1)std(l_j(x^{(m)}_{j}))}{\sum_{m=1}^M(N_m-1)}},
\end{align*}
where $std\left(\cdot\right)$ is the standard deviation of the training data.

\noindent{\textbf{Variable importance}} Variable importance is a widely used approach for the post-hoc interpretation of black-box ML models. It provides insights into the ranking of variables based on their contributions to the outcomes. Variable importance was computed based on the importance of the rule terms and the importance of the linear terms as follows:
\begin{align*}
\displaystyle
I_{j}(\bm{x}) = I_j(x_{j}) + \sum_{x_{j}\in r_k} \frac{I_k(\bm{x})}{m^*_k},
\end{align*}
where the first term $I_j(x_j)$ denotes the importance of the $j$th linear term, the second term denotes the sum of the importance of the rules with $x_j \ (x_j\in r_k)$, and $m^*_k$ is the total number of variables $x_{j}$ used to define the rule.

\section{Numerical simulation}\label{sec3}
In this section, we described a series of numerical simulations conducted to evaluate the predictive performance of the proposed federated RuleFit approach. The proposed method was compared with that of centralized RuleFit and several existing FL methods. To provide a comprehensive evaluation, we considered multiple data generation models and several evaluation metrics for each challenge scenario. First, we described the simulation design. Then, we presented the simulation results for the different experimental settings.

\subsection{Simulation design}

In this simulation study, we considered an FL setting with $M$ clients. For each simulation setting, we first generated a global training dataset and test dataset of the form $\{(y_i, \bm{x}_i)\}_{i=1}^{N}$. Both datasets were partitioned into $M$ disjoint subsets of size $N_1, \ldots, N_M$, denoted by $\{(y^{(m)}_i, \bm{x}^{(m)}_i)\}_{i=1}^{N_m}$ for $m = 1, 2, \ldots, M$, to simulate a federated setting, where each subset corresponds to the local dataset of the client.

The covariate vector $\bm{x}_i = (x_{i1}, x_{i2}, \ldots, x_{i10})^{\top}$ was independently generated from the multivariate normal distribution $\mathcal{N}(\bm{0}, \bm{\Sigma})$, where $\bm{\Sigma}$ is a $10 \times 10$ diagonal covariance matrix with unit variances. The first five covariates were informative, whereas the remaining were noise variables.

We considered both linear and nonlinear relationships between the covariates and outcomes as follows:
\begin{align*}
\mathrm{Model\ 1}: \quad \eta_i &= 5x_{i1} - 4x_{i2} + 3x_{i3} - 2x_{i4} + x_{i5},\\
\mathrm{Model\ 2}: \quad \eta_i &= 10\exp(-2x_{i1}^2) - 10\exp(-2x_{i2}^2) \\
&+ 6\sin(x_{i3}) - 4\sin(x_{i4}) + 2\sin(x_{i5}) 
\end{align*}

Outcomes were generated according to the following logistic model:
\begin{align*}
y_i \sim \mathrm{Bernoulli}\left( \frac{\exp(\eta_i)}{1 + \exp(\eta_i)} \right).
\end{align*}

We considered three scenarios in FL: (1) different numbers of participating clients, (2) client-level sample size imbalance, and (3) outcome imbalance across clients, as discussed in \cite{Hauschild2022}. For each scenario, we fixed the total sample size aggregated from all the institutions at $N = N_1 + N_2 + \cdots + N_M = 1000$.

\begin{itemize}

\item \textbf{Scenario 1: Different numbers of participating clients} 
In this scenario, we examined how the number of participating clients affects predictive performance under a fixed total sample size. We considered four settings with different numbers of clients, namely $M = 2, 5, 10$, and $20$. The total sample size was fixed at $N = 1000$, and the data were evenly partitioned across clients, ensuring that $N_m = N / M$ for all $m = 1, 2, \ldots, M$. This setup allowed us to evaluate the impact of data fragmentation as the number of clients increases and local sample sizes decrease.

\item \textbf{Scenario 2: Client-level sample size imbalance}
In this scenario, we investigated how imbalanced data availability across clients affects predictive performance. We fixed the total sample size at $N = 1000$ and the number of clients at $M = 5$. Three configurations of client-level sample proportions were considered: balanced ($0.2, 0.2, 0.2, 0.2, 0.2$), moderately imbalanced ($0.1, 0.15, 0.2, 0.25, 0.3$), and highly imbalanced ($0.05, 0.1, 0.15, 0.25, 0.45$). This setup allowed us to evaluate the effect of client-level sample size imbalance on model performance in federated settings.

\item \textbf{Scenario 3: Outcome imbalance across clients}
In this scenario, we investigated how variation in outcome prevalences across clients affects predictive performance. We fixed the total sample size at $N = 1000$ and the number of clients at $M = 5$ and evenly partitioned the data across clients, ensuring that $N_m = N / M$ for all $m = 1, 2, \ldots, M$. We then altered the client-specific prevalence of the positive class. Three configurations are considered: balanced ($0.5, 0.5, 0.5, 0.5, 0.5$), moderately imbalanced ($0.25, 0.375, 0.5, 0.625, 0.75$), and highly imbalanced ($0.125, 0.250, 0.375, 0.875, 0.875$). This setup allowed us to assess the impact of outcome imbalance on model performance in federated settings.

\end{itemize}

We evaluated the performance of the proposed approach using multiple metrics and compared it with the following baseline methods.

\begin{itemize}

\item \textbf{Centralized RuleFit:} 
A RuleFit model was trained using pooled data from all clients at a central server. This represented the conventional RuleFit model without the constraints of FL and served as a reference benchmark for assessing how closely the proposed federated approach approaches the centralized setting. Furthermore, since the total sample size and data-generating process were fixed within each scenario, the performance of the centralized RuleFit remained largely unchanged across different settings; therefore, it was regarded as a reference benchmark.

\item \textbf{Local RuleFit:} 
A RuleFit model was trained independently by each client using only local data, representing a non-collaborative setting without FL. The performance was evaluated through averaging the results across clients. This baseline was used to explicitly quantify the performance gain achieved by extending RuleFit to FL settings.

\item \textbf{Bayesian Federated Inference (BFI):} 
An FL approach was based on a Bayesian framework that constructs a global linear model through aggregating local posterior distributions without sharing raw data \cite{Jonker2024}. This method served as a linear baseline in the federated setting. An implementation is available via the \textit{BFI} R package.

\item \textbf{Federated Random Forest (FRF):} 
An FL method trained Random Forest models locally by each client and aggregated them to construct a global model \cite{Hauschild2022, Herrera2024}. This method represented a nonlinear ensemble baseline based on local model aggregation and was included as a relevant comparison, as it was based on local tree-model training for each client followed by model aggregation, which was conceptually similar to the local rule generation stage of the proposed method. The method was implemented using the \textit{flextrees} Python package.
\end{itemize}

All methods required the specification of hyperparameters prior to model training. For the centralized and local RuleFit models, we adopted the parameter settings recommended by Friedman and Popescu \cite{Friedman2008}, where the number of base learners, mean depth of each base learner, and shrinkage rate were set to 333, 2, and 0.01, respectively. For the proposed method, the same hyperparameter settings were used in the rule-generation stage to ensure comparability with the baseline methods. In the preprocessing step, the privacy budget for each DP histogram was set to $\epsilon = 1$, which provided a reasonable balance between privacy protection and data utility. In the rule ensemble stage, the hyperparameters of the FedDA algorithm were selected according to \cite{Yuan2021}, where similar parameter settings were used for federated Lasso problems. Specifically, the server learning rate was set to $\eta_s = 1$, client learning rate to $\eta_c = 0.01$, number of server updates to $R = 300$, and number of client updates to $G = 20$. The regularization parameter was set to $\lambda = 0.01$ to promote sparsity in the model and improve interpretability. For BFI and FRF, we used their default parameters.

\subsection{Results}

The results are reported as differences relative to the centralized RuleFit, the performance of which is stable across all scenarios. Because all evaluation metrics [area under the receiver operating characteristic curve (AUC), accuracy, and F1-score] were higher, positive values indicated better performance than the centralized RuleFit, whereas negative values indicated worse performance. Detailed results for each metric are provided in the Appendix 3.

\paragraph{Overall performance}
Across all scenarios, the proposed method consistently achieved a predictive performance comparable to that of the centralized RuleFit. In contrast, the local RuleFit exhibited substantial performance degradation, particularly when local sample sizes were small or the outcome distributions were imbalanced across clients. Compared to other federated methods, the proposed approach consistently outperformed FRF. Although BFI performed best under linear data-generating processes owing to its linear modeling structure, the proposed method remained competitive in such settings, achieving a higher predictive accuracy in nonlinear scenarios.

\paragraph{Scenario 1: Number of clients}
Figure~2(A) shows the results as the number of clients increased from $M=2$ to $M=20$, corresponding to increasingly severe data fragmentation and smaller local sample sizes. Under this setting, the proposed method consistently achieved a performance close to that of the centralized RuleFit across all metrics, with differences remaining near zero and showing minimal variability, indicating strong robustness to data fragmentation. Within the RuleFit framework, a clear contrast was observed; whereas the proposed method remained stable, the local RuleFit exhibited a monotonic decline as $M$ increased, indicating that local models failed to capture the relationships between the outcome and covariates under limited data. Compared to other federated methods, the proposed approach outperformed FRF across all settings. Although BFI achieved competitive performance in the linear setting, its performance deteriorated in the nonlinear one, whereas the proposed method remained stable across both settings.

\paragraph{Scenario 2: Sample size imbalance}
Figure~2(B) shows the results of a client-level sample size imbalance, where the amount of available data varied across clients. The proposed method maintained a performance equivalent to that of the centralized RuleFit across all imbalance configurations, with only minor deviations and limited variability, indicating its robustness to imbalanced sample sizes across clients. However, the local RuleFit was more sensitive to imbalances because clients with smaller sample sizes produced less reliable local models, leading to a noticeable decline in performance. In contrast, the proposed method effectively integrated information across clients and mitigated this issue. Compared with other federated methods, the proposed approach outperformed FRF. BFI remained competitive in the linear setting; however, its advantage diminished in the nonlinear one, where the proposed method provided a more consistent performance.

\paragraph{Scenario 3: Outcome imbalance}
Figure~2(C) shows the results for outcome imbalance across clients, representing a challenging setting in which institutions have substantially imbalanced outcomes. In this scenario, the proposed method remained stable and achieved a performance comparable to that of the centralized RuleFit across all metrics, demonstrating strong robustness to heterogeneity in outcome prevalence. However, the local RuleFit exhibited pronounced performance degradation, particularly in nonlinear settings, because models trained on imbalanced local data were strongly influenced by the skewed outcome distribution for each client. In contrast, the proposed method effectively aggregated information across clients and alleviated this issue. Compared with other federated methods, the proposed approach outperformed FRF, whereas BFI performed well in the linear setting but showed reduced prediction accuracy in the nonlinear one. Overall, the proposed method provided a more consistent performance across various outcome distributions.

\begin{figure}[tb]
  \centering
  \includegraphics[width=\linewidth]{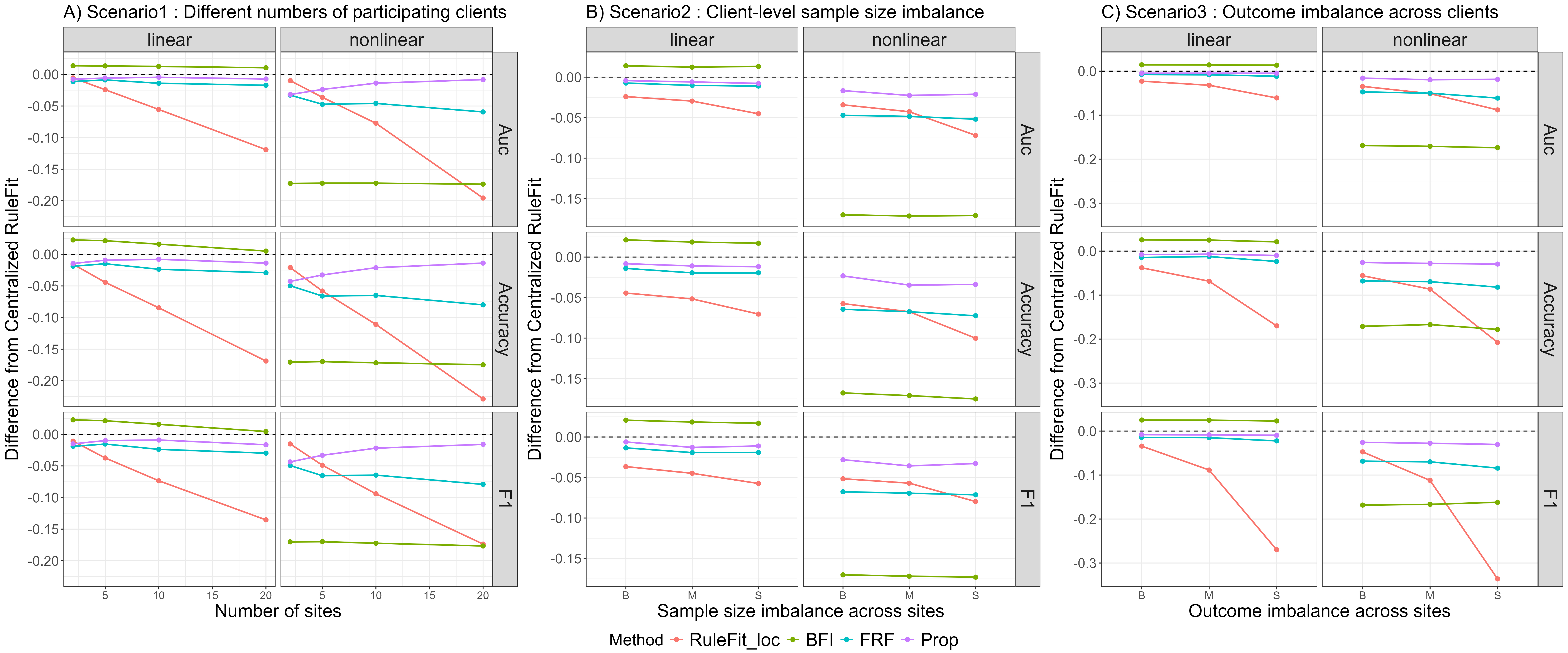}
  \caption{Predictive performance under three simulation scenarios, expressed as differences relative to the proposed method. Positive values indicate a better performance than that of the proposed method, whereas negative values indicate worse performance. The horizontal dashed line at zero represents no performance differences compared to the proposed method. Panel A corresponds to Scenario 1, where the number of participating clients varies ($M = 2, 5, 10, 20$); Panel B corresponds to Scenario 2, where client-level sample sizes are imbalanced (B: balanced, M: moderately imbalanced, and S: substantially imbalanced); and Panel C corresponds to Scenario 3, where the prevalence of the positive outcome varies across clients. Results are shown separately for the linear and nonlinear data-generating processes.}
  \label{Fig2}
\end{figure}

\section{Real-world data application}\label{sec4}
In this section, we applied the proposed approach to a trauma dataset, originally reported in \cite{Draaisma1989} and later analyzed in \cite{Jonker2024}, to evaluate its performance on real-world data. The dataset comprised the information of 371 patients from three different institutions: a peripheral hospital without a neurosurgical unit (NSU), a peripheral hospital with an NSU, and an academic medical center. The sample sizes for these institutions were 49, 106, and 216, respectively. The outcome variable of interest was binary mortality. The covariates included age, sex (0: male, 1: female), Injury Severity Score (ISS), and Glasgow Coma Scale (GCS). The ISS is a widely used clinical score for assessing trauma severity, ranging from 1 to 75, with higher values indicating more severe injury. The GCS is a clinical scale used to evaluate a patient’s level of consciousness, ranging from 3 to 15, where higher values indicate better neurological status.

\subsection{Application results}

To assess the application of the proposed method and illustrate its clinical interpretation, the hyperparameters were set as those in the simulations. The model generated 82 rules, each characterized by importance, support, and coefficient. Importance reflects the contribution of a rule to the outcome, support represents the proportion of participants satisfying the rule, and the coefficient indicates the direction and magnitude of the rule’s effect. To enhance interpretability and robustness, we focused on rules that were both important and sufficiently supported by selecting the top five rules with support greater than 0.1.

\begin{table}[tb]
\caption{Top five most important rules with support greater than 0.1}
\centering
\label{Real_data_table}
\begin{tabular}{l|lccc}
\hline
No.    & Rules                                                                   & exp(coef.) & Rule importance & Support \\ \hline
Rule 1 & age \textless 57.23 \& GCS \textgreater{}= 5.086 \& ISS \textless 57.66 & 0.80       & 100             & 0.50    \\
Rule 2 & age \textgreater{}= 22.27 \& ISS \textgreater{}= 26.73                  & 1.21       & 83              & 0.50    \\
Rule 3 & age \textless 56.64 \& ISS \textless 57.66                              & 0.82       & 82              & 0.68    \\
Rule 4 & age \textless 57.23 \& ISS \textless 57.66                              & 0.82       & 80              & 0.69    \\
Rule 5 & age \textless 33.4 \& GCS \textgreater{}= 3.703 \& ISS \textless 57.66  & 0.84       & 79              & 0.43    \\ \hline
\end{tabular}
\end{table}

Table~\ref{Real_data_table} summarizes the selected rules and their corresponding effects. Overall, rules involving low GCS values and high ISSs were associated with an increased mortality risk, whereas those involving higher GCS values were associated with a reduced mortality risk. For example, patients satisfying Rule2 exhibited higher mortality rates (exp(coef.) = 1.21), indicating increased odds of mortality, whereas rules with exp(coef.) corresponded to protective patterns. These findings align with established clinical knowledge, where higher GCS scores reflect better neurological status and a higher ISS indicates more severe trauma.

\begin{figure}[!t]
  \centering
  \includegraphics[width=\linewidth]{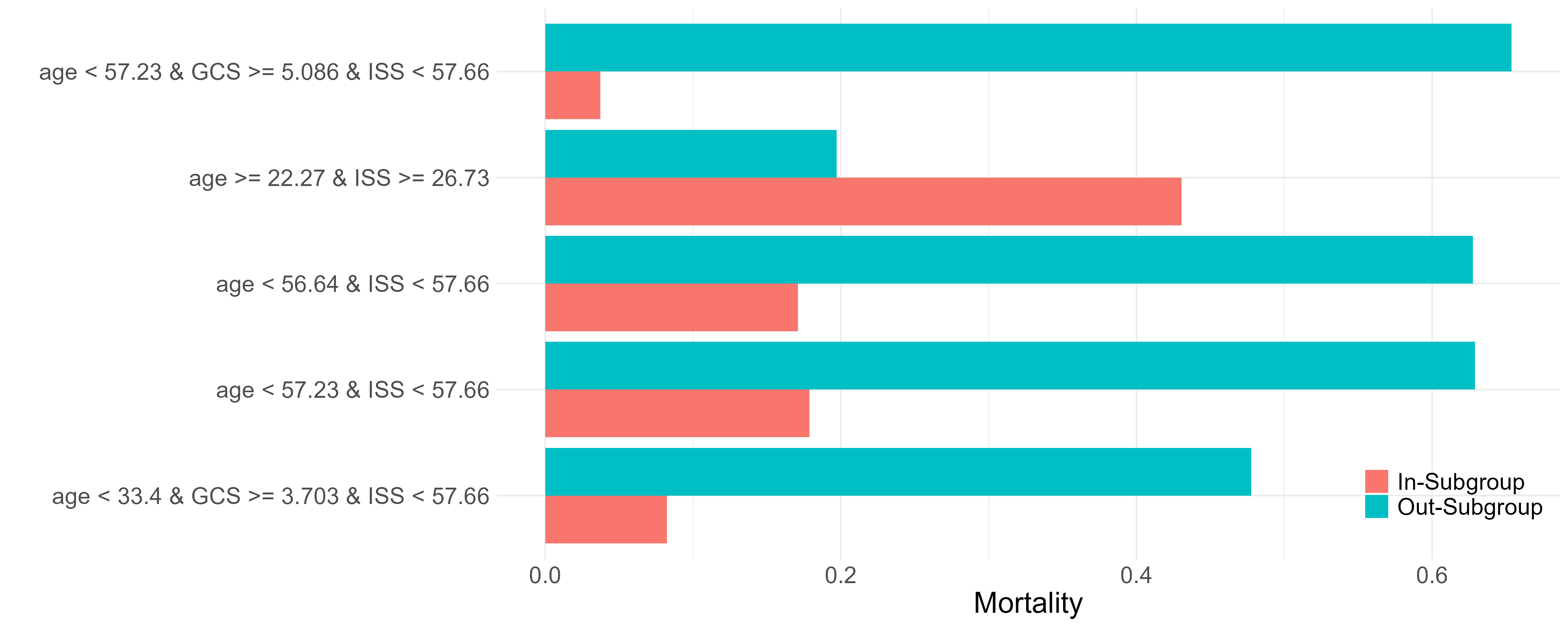}
  \caption{Subgroup evaluation for each rule. Bars represent mortality rates within (in-subgroup) and outside (out-of-subgroup) the population defined by each rule.}
  \label{Fig3}
\end{figure}

We further evaluated mortality within the subgroups defined by each rule and compared it to the corresponding out-of-subgroup mortality (Figure~\ref{Fig3}). A clear separation between in-subgroup and out-of-subgroup mortality was observed across all rules, with particularly large differences in rules involving low GCS scores. This indicates that the extracted rules effectively stratified patients into clinically distinct risk groups.

\subsection{Performance evaluation}
For the real-data application, data from each institution were randomly split into training and test sets, with 70\% used for training and 30\% used for testing. This division was conducted independently within each institution to preserve the federal setting. The model was trained using training sets from all the institutions under the federated framework.

\begin{figure}[!t]
  \centering
  \includegraphics[width=\linewidth]{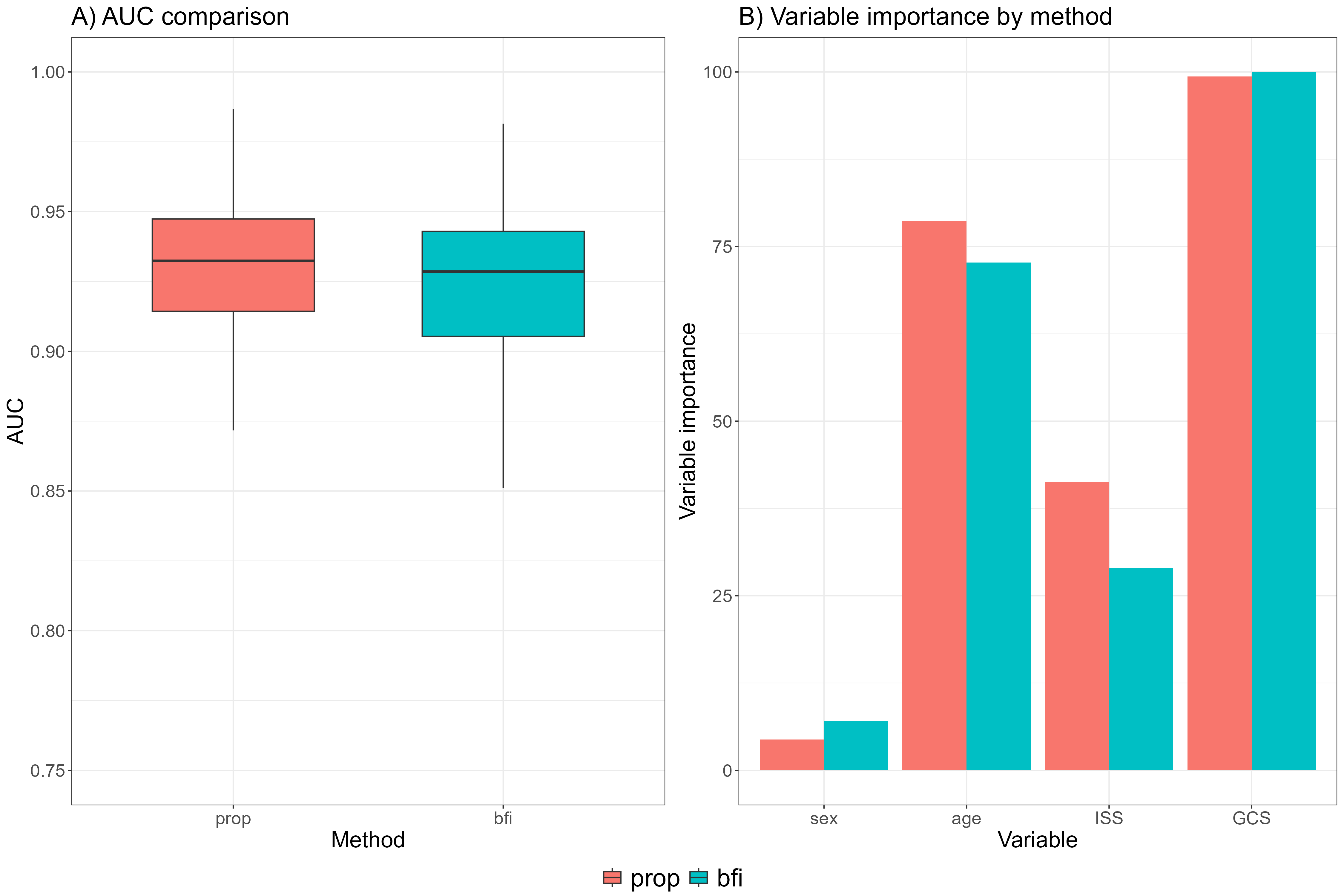}
  \caption{Subgroup evaluation for each rule}
  \label{Fig4}
\end{figure}

We first evaluated the predictive performance of the proposed method using test sets and compared it with BFI, which was previously applied to analyze this dataset \cite{Jonker2024}. The predictive performances are summarized in Fig\ref{Fig4}. The boxplot of the AUC shows that the proposed method achieved competitive predictive performance compared to BFI. In particular, the proposed method exhibited slightly higher median AUC values with relatively stable variability across repetitions, indicating its robust predictive ability.

We also examined the importance of each method. As shown in Fig\ref{Fig4}, both methods consistently identified GCS as the most influential variable, followed by age and ISS, whereas sex has comparatively minor importance. Notably, the proposed method assigned relatively higher importance to age and ISS than to BFI, suggesting an enhanced ability to capture clinically relevant covariates and their interactions.

Overall, these results indicate that the proposed method achieved comparable or improved predictive performance relative to BFI, while providing more interpretable insights into variable importance. This supports the effectiveness of the proposed approach for prediction and clinical interpretability in a FL setting.

\section{Conclusion and Discussion}\label{sec5}

We proposed a federated RuleFit framework for constructing a unified and interpretable global model without sharing raw data across clients. The proposed approach achieved a predictive performance comparable to RuleFit in a centralized setting while preserving interpretability, indicating that the tradeoff between interpretability and predictive accuracy can be effectively mitigated in federated settings.

The results of our numerical study (Section 3) highlighted the fundamental limitation of naïvely applying RuleFit to federated settings. Independently trained local RuleFit models are adversely affected by limited sample sizes and outcome imbalances across clients, leading to a degraded predictive performance and inconsistent model structures. These models capture only client-specific relationships between the covariates and outcomes, reflecting the limited and varying data available at each site. Moreover, because the models are constructed independently, their rule structures and selected features vary substantially across clients; therefore, they fail to provide a unified and interpretable global representation.

To address these limitations, we adopted a hybrid strategy that combined local rule generation with federated aggregation and model estimation, enabling the construction of a unified and interpretable global model from locally derived rules. However, heterogeneity in local data distributions induces variations in the split points, leading to an expansion of the candidate rule set and increased computational burden. To overcome this challenge, we introduced a preprocessing step based on DP histograms to construct a shared set of candidate cut-off values. This component harmonized the split points across clients in a privacy-preserving manner and constrained the growth of the candidate rule set. The results presented in Appendix 2 show that using a relatively small number of global cutoff values (e.g., 20) achieved a predictive performance comparable to that obtained without the preprocessing step, while substantially reducing both the computational cost and the number of rule terms in the final model.

The numerical simulation results further demonstrated that the proposed method was robust across various scenarios, including an increasing number of clients, sample size imbalances, and outcome imbalances. Compared with other federated methods, the proposed approach consistently outperformed FRF, indicating that the simple aggregation of locally trained models is insufficient under heterogeneous data conditions. Although BFI performed well in linear settings, its performance deteriorated in nonlinear scenarios, whereas the proposed method maintained a stable performance across both settings.

The proposed method was further evaluated using a real-world trauma dataset originally analyzed by Jonker et al.\cite{Jonker2024}, which facilitated direct comparison under identical data conditions. The results showed that the proposed method yielded clinically meaningful and interpretable insights, while achieving a slightly higher predictive accuracy than that of BFI. Moreover, the consistency in variable importance between both methods supports the reliability of the interpretations. Collectively, these findings indicate that the proposed approach is effective for constructing accurate and interpretable models in FL settings.

In conclusion, the proposed federated RuleFit framework achieves both strong predictive performance and interpretability. However, several limitations should be noted. First, this study focuses on binary outcomes, whereas extensions to continuous and survival outcomes are important for broader applicability in medical research. Second, the current framework is limited to horizontal federated learning; extensions to vertical or hybrid settings remain for future work.

\bibliographystyle{unsrt}
\bibliography{references}

\clearpage
\appendix
\section{Differential Privacy and Laplace Mechanism}\label{A1}

\subsection{Differential Privacy} 
Differential privacy is a mathematically rigorous framework for protecting individual-level data \cite{Dwork2006}. It guarantees that the outcome of a statistical analysis remains nearly unchanged whether or not any single individual's data is included in the dataset.

Formally, given any two neighboring datasets $D$ and $D'$ that differ in at most one record, a randomized mechanism $\mathcal{F}$ satisfies $\epsilon$-differential privacy if for all measurable sets $S$,
\begin{align*}
\Pr[\mathcal{F}(D) \in S] 
\leq 
\exp(\epsilon)\Pr[\mathcal{F}(D') \in S],
\end{align*}
where $\epsilon > 0$ is the privacy budget. A smaller $\epsilon$ corresponds to a stronger privacy guarantee.

\subsection{Laplace Mechanism}
Given a query function $f: \mathcal{D} \rightarrow \mathbb{R}^d$ with global sensitivity
\begin{align*}
\Delta = \max_{D,D'} \| f(D) - f(D') \|_1,
\end{align*}
where $D$ and $D'$ are neighboring datasets that differ in at most one record, 
the Laplace mechanism defines a randomized mechanism $\mathcal{M}$ as
\begin{align*}
\mathcal{M}(D) = f(D) + \bm{\eta},
\end{align*}
where each component of $\bm{\eta} \in \mathbb{R}^d$ is independently drawn from
\begin{align*}
\mathrm{Lap}\!\left(0, \frac{\Delta}{\epsilon}\right).
\end{align*}
Then $\mathcal{M}$ satisfies $\epsilon$-differential privacy.

\clearpage
\section{Impact of the number of bins and quantiles selection in pre-process on the predictive performance}\label{A2}

\begin{figure}[ht]
  \centering
  \includegraphics[width=\linewidth]{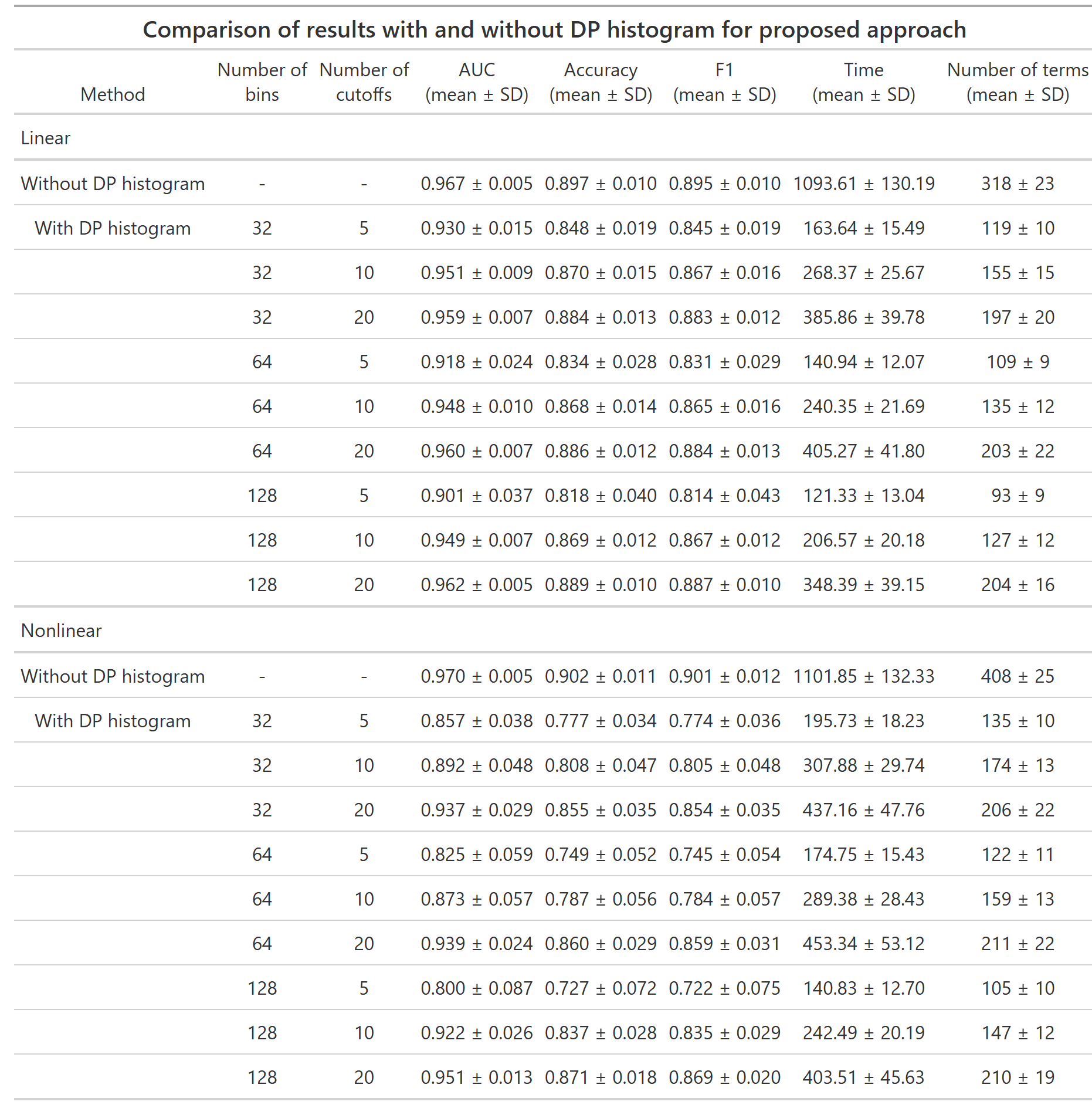}
  \caption{Impact of preprocessing design choices on predictive performance for the proposed method. The number of bins and quantile cut-offs used in the differentially private (DP) histogram are varied under Scenario 1, with the number of clients fixed at $M=5$. Performance is evaluated in terms of AUC, accuracy, F1-score, computation time, and the number of extracted rules. Results are compared with a baseline without DP histogram. Both linear and nonlinear data-generating processes are considered.}
  \label{Fig6}
\end{figure}

\clearpage
\section{Detailed Results for Each Metric}\label{A3}
\begin{figure}[ht]
  \centering
  \includegraphics[width=0.95\linewidth]{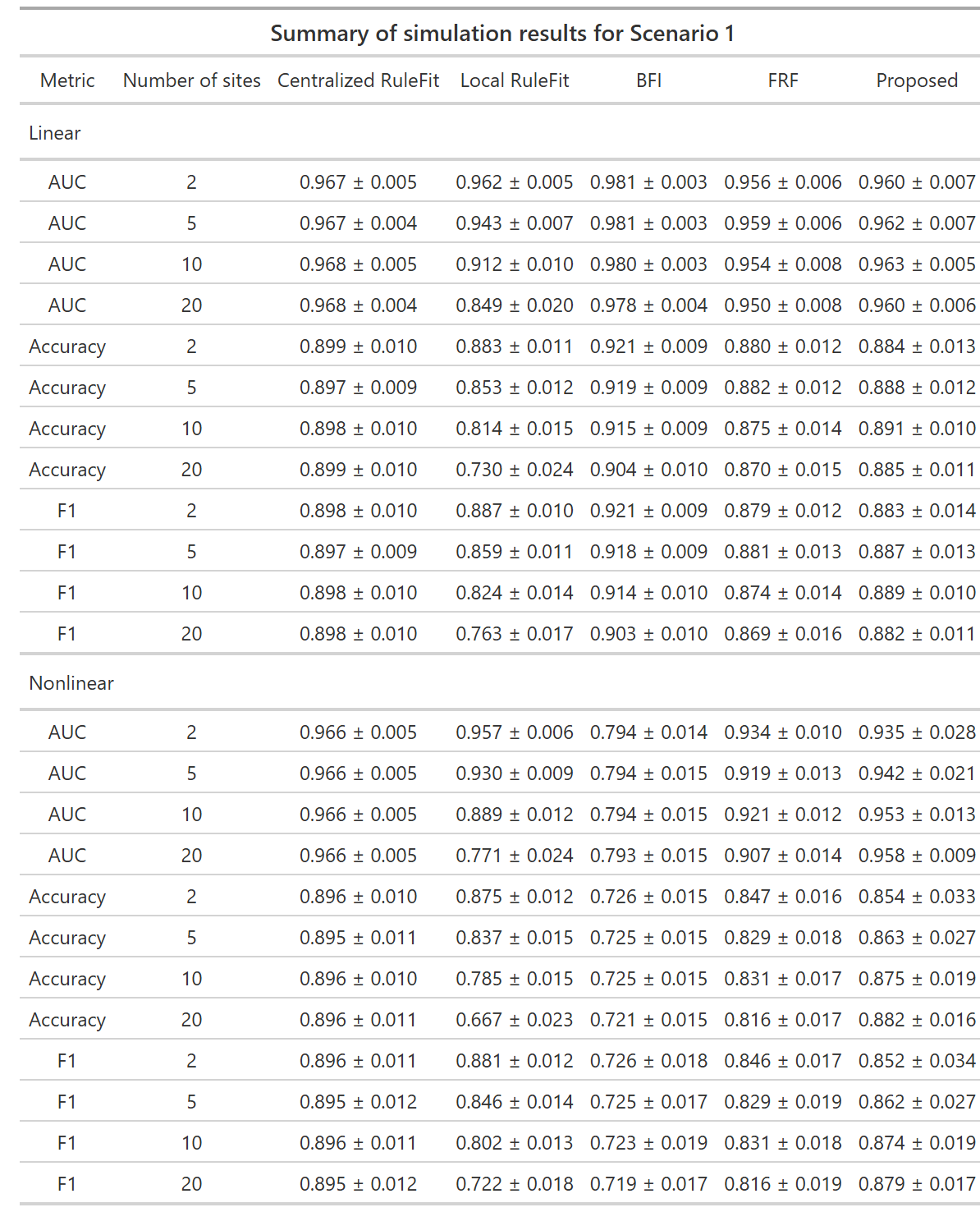}
\end{figure}

\begin{figure}[ht]
  \centering
  \includegraphics[width=0.95\linewidth]{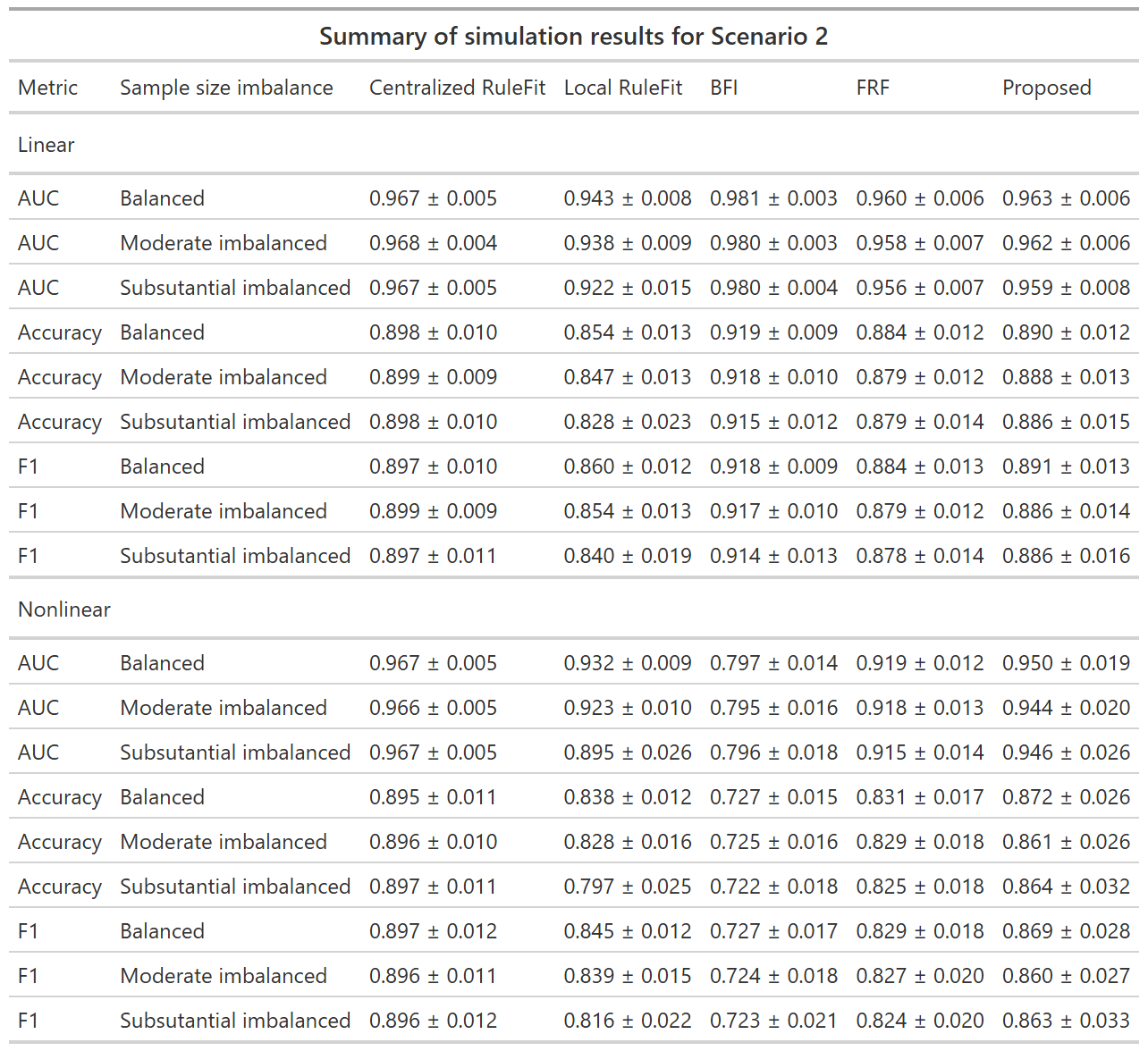}
\end{figure}

\begin{figure}[ht]
  \centering
  \includegraphics[width=0.95\linewidth]{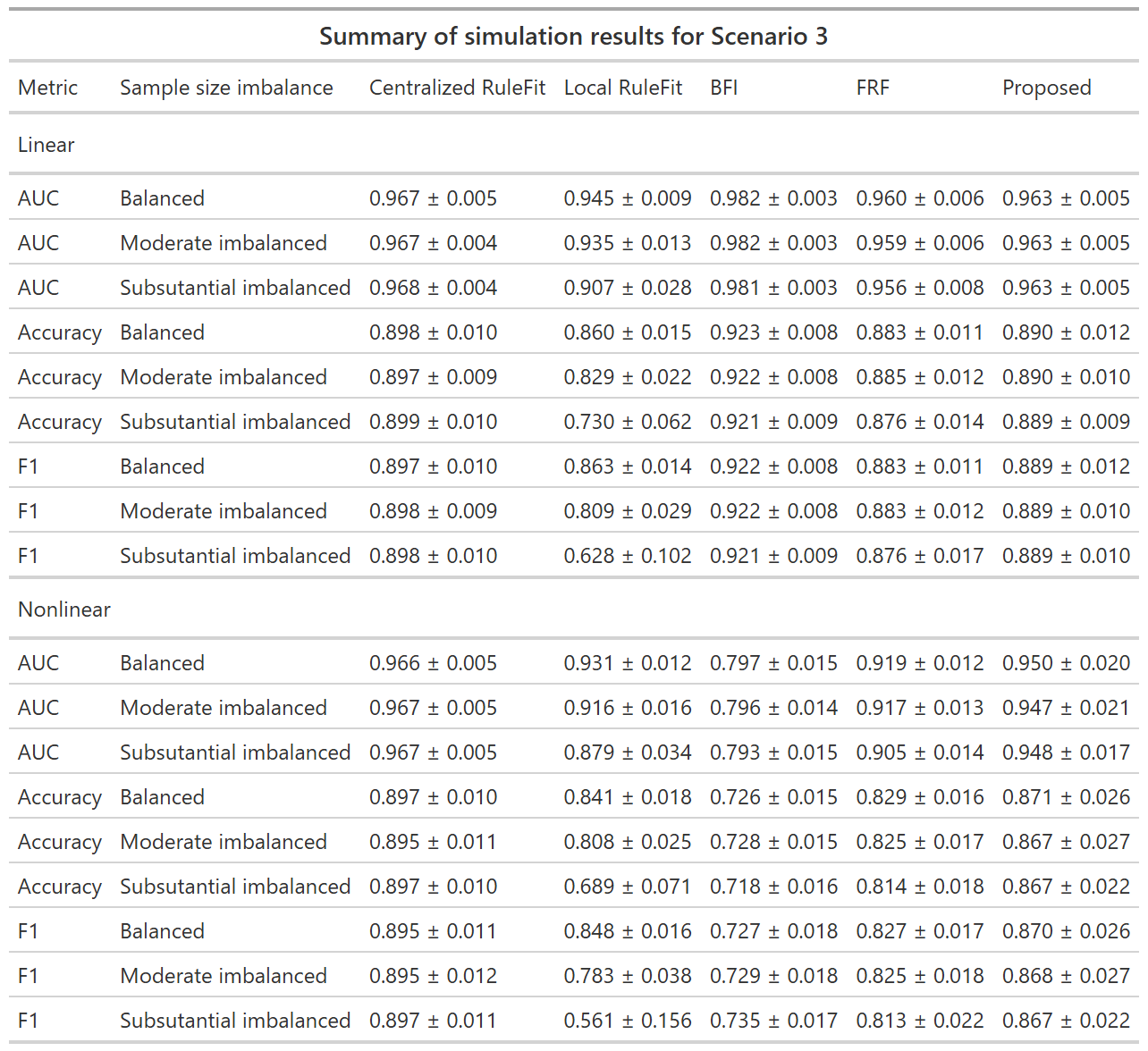}
\end{figure}

\end{document}